\def\year{2018}\relax
\pdfoutput=1
\documentclass[letterpaper]{article} 
\usepackage{aaai18}  
\usepackage{subcaption}
\usepackage{multirow}
\usepackage{times}  
\usepackage{helvet}  
\usepackage{courier}  
\usepackage{url}  
\usepackage{graphicx}  
\begin{document}
%
\title{Image Captioning at Will: A Versatile Scheme for Effectively Injecting Sentiments into Image Descriptions}
\author{Quanzeng You\\
	Microsoft \\
	One Microsoft Way \\
	Redmond, WA 98052 \\
	quanzeng.you@microsoft.com \\
	\And Hailin Jin \\
	Adobe Research \\
	345 Park Avenue\\
	San Jose, CA 95110\\
	hljin@adobe.com \\
	\And Jiebo Luo\\
	Department of Computer Science \\
	University of Rochester \\
	Rochester, NY 14627 \\
	jluo@cs.rochester.edu \\
}
\maketitle
\begin{abstract}
Automatic image captioning has recently approached human-level performance due to the latest advances in  computer vision and natural language understanding. However, most of the current models can only generate plain factual descriptions about the content of a given image. However, for human beings, image caption writing is quite flexible and diverse, where additional language dimensions, such as emotion, humor and language styles, are often incorporated to produce diverse, emotional, or appealing captions. In particular, we are interested in generating sentiment-conveying image descriptions, which has received little attention. The main challenge is how to effectively inject sentiments into the generated captions without altering the semantic matching between the visual content and the generated descriptions. In this work, we propose two different models, which employ different schemes for injecting sentiments into image captions. Compared with the few existing approaches, the proposed models are much simpler and yet more effective. The experimental results show that our model outperform the state-of-the-art models in generating sentimental (i.e., sentiment-bearing) image captions. In addition, we can also easily manipulate the model by assigning different sentiments to the testing image to generate captions with the  corresponding sentiments.
\end{abstract}
	\section{Introduction}
\label{sec:intro}
Recently, automatic image captioning has attracted a lot of attention in both vision and language research communities. This is not only due to its potential applications, such as helping visually impaired people. More importantly, the advances made in solving this problem will further bridge the semantic connection between natural language processing and computer vision. It is worth noting that a large number of recent developments in both areas follow the success of deep neural networks. Different novel and effective deep neural architectures have been proposed to improve the performance of current image captioning systems.  On the other hand,  further advances in image captioning can in turn benefit the research in both computer vision and natural language processing.

Currently, the majority of the research efforts have been trying to produce human-level factual descriptions for a given image. Indeed, this is related to several core computer vision tasks, including image classification~\cite{haralick1973textural,yang2009linear} and object detection~\cite{felzenszwalb2010object,viola2001rapid}, which attempt to understand the objects, attributes and factual content in images. Generating factual descriptions of the images is also challenging and interesting in terms of understanding the visual content and producing semantically matching and grammatically correct sentences. However, compared with manually written captions, which will consider various dimensions, such as context, personal feelings and language styles,  current machine generated factual captions are always plain and boring. In this work, we investigate the injection of sentiments into the generated image captions because sentiment is one of the rich dimensions in languages as discussed in~\cite{leech2016principles,mathews2016senticap}.
\begin{figure}
	\centering
	\includegraphics[width=0.4\textwidth]{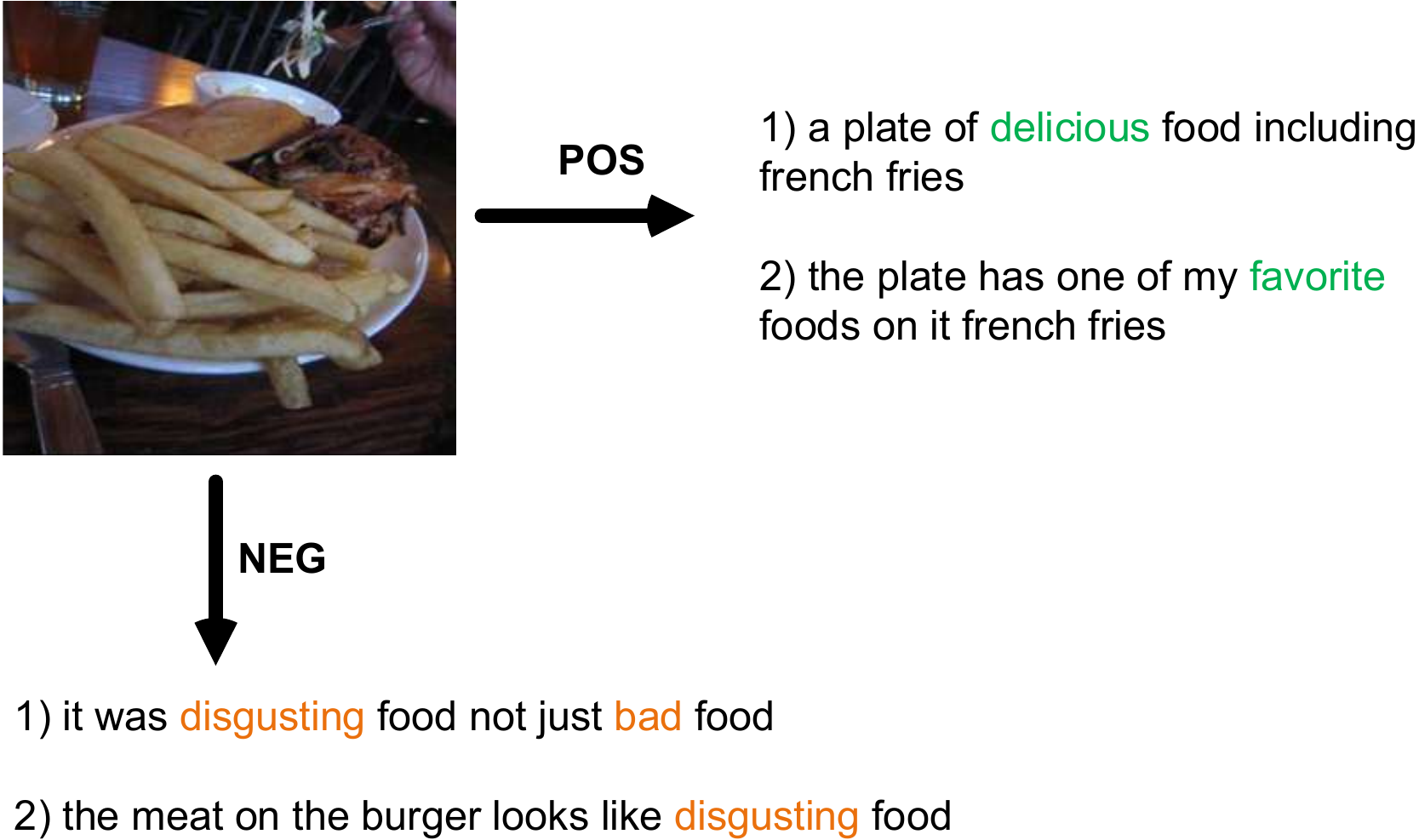}
	\caption{An example of injecting sentiment. Two research problems: 1) injection of sentiments, where the sentiment related words are highlighted in colors; and 2) controllable injection of sentiments, where POS and NEG stands for positive and negative sentiments, respectively.}
	\label{fig:motivation}
	\vspace{-10pt}
\end{figure}

\figurename~\ref{fig:motivation} illustrates the two problems we intend to solve in this work. The first problem is to include the sentiment dimension in the produced captions. Equivalently, the problem is to inject sentiments into the captions, while trying to preserve the semantic mapping between the images and the captions. The example captions in \figurename~\ref{fig:motivation} contain highlighted sentiment words in colors. The second problem is being able to control the injection of different emotions. Specifically, we can inject either positive or negative sentiment into the caption by providing different sentiment labels as we wish. 

To solve these two problems, we design two novel mechanisms targeting on the injection of sentiment into the generation of captions. The first proposed approach \textit{directly} supplies the sentiment label as one additional dimension of the input feature to each step of the language generation model, which is a recurrent neural network (RNN). In such a way, we expect the additional sentiment dimension, which can be considered as the equivalent of a bias term in the model, will learn to control the injection of sentiment words in generating captions. At each step, this additional dimension provides extra information for the model to rerank the words for the next step. The second proposed approach intends to propagate the sentiment information, which is provided at the initial step of the LSTM model, over the whole sequence. Inspired by the memory cell in the LSTM model, we devise a sentiment cell to propagate the sentiment signal, which employs a similar computational procedure to the memory cell. Both approaches can be trained in an end-to-end fashion for building an automatic sentiment captioning system. 

Meanwhile, the sentiment signal can be easily supplied to the model in both approaches. Therefore, we can easily control the generation of the captions by providing different sentiment labels. In such a way, we can easily produce different sentiment captions. Compared with the state-of-the-art approaches on sentiment image captioning, the proposed models achieve better performance in terms of benchmark evaluation metrics. Further experimental analysis also suggests that the two designed sentiment units can help the model to distinguish different sentiments and thus generate the corresponding sentiment related captions. We make the following contributions in this work:
\begin{itemize}
	\item We design two different models for injecting sentiments into the generated captions. Both models can be trained in an end-to-end fashion.
	\item The sentiment signals are supplied in two simple yet effective ways so that we can easily produce a desired caption for a given image and a given sentiment.
	\item Our approaches significantly outperform the state-of-the-art approaches in generating sentimental captions.
\end{itemize}


\section{Related Work}
\label{sec:related}
Our work is mostly related to the current research on image captioning and text generation. In this section, we review the most relevant studies on these two topics.
\subsection{Image Captioning}
Recent studies on image captioning have been focusing on the application of deep neural networks since the release of MS-COCO Image Captioning Challenge\footnote{\scriptsize\url{http://mscoco.org/dataset/#captions-challenge2015}}. The authors from~\cite{lebret2015phrase,fang2015captions} first apply deep learning to predicting words and phrases from the  given images. The captions are then generated by another language model, which composes the candidate words into a sentence.

Meanwhile, most of recent publications have been using an encoder-decoder framework~\cite{cho2014learning,sutskever2014sequence} for decoding the encoded images into a sentence. The work in~\cite{vinyals2015show} proposed a CNN-RNN framework, which is simple but very effective. Their model ranked first in the 2015 MS-COCO Image Captioning Challenge. Both~\cite{Mao_2015_ICCV} and \cite{mao2014deep} employed the multimodal RNN for learning the semantic mapping between images and words, where the encoded image is supplied at each step of the RNN for learning the multimodal layer. More recently, attention model~\cite{bahdanau2014neural,xu2015show,you2016image,liu2016attention,lu2016knowing}, which tries to learn the alignment between source language and target language in machine translation, has widely been adopted for building better image captioning systems. 
\subsection{Text Generation}
Image captioning is also a text generation process. The uniqueness comes from the input, which is the encoded image. Currently, recurrent neural network has the state-of-the-art performance on text generation~\cite{bowman2015generating,ha2016hypernetworks}. However, Variational Autoencoders (VAE)~\cite{kingma2013auto,rezende2014stochastic} have been widely studied as a generative model. Specifically, it has been applied to text generation~\cite{bowman2016generating,semeniuta2017hybrid}, where the main challenges come from the collapse of the latent loss. Both studies have proposed different mechanisms to solve this issue. More recently, \cite{hu2017controllable} offered another variant of VAE for text generation. They have explicitly included structured attributes to the hidden representation of the sentence produced by the encoder. Indeed, this is close to our proposed direct injection of sentiments. However, we have employed the attributes in a step-wise style, where they are only appended to the attributes as part of the global encoded representation.

In summary, generating different styles of captions requires us to not only bridge the semantic meanings between image and text, but also build a language generative model that can understand the differences between different language styles. This makes it more difficult for building sentimental captioning systems. There have been several preliminary studies~\cite{mathews2016senticap,shinimage,karayil2016generating}. However, ~\cite{mathews2016senticap} proposed a model that cannot be trained in an end-to-end fashion. There were only some preliminary example results in~\cite{karayil2016generating}. On the other hand, ~\cite{shinimage} tried to learn the model from large scale weakly and noisily supervised data, where additional care needs to be taken to reduce the noises. In contrast, our proposed approach is simple to train, yet it significantly outperforms the state-of-the-art.
\section{Models for Injection of Sentiments}
\label{sec:model}
In this section, we introduce the proposed approaches for injection of sentiments into the generated captions.
\begin{figure}
	\includegraphics[width=0.475\textwidth]{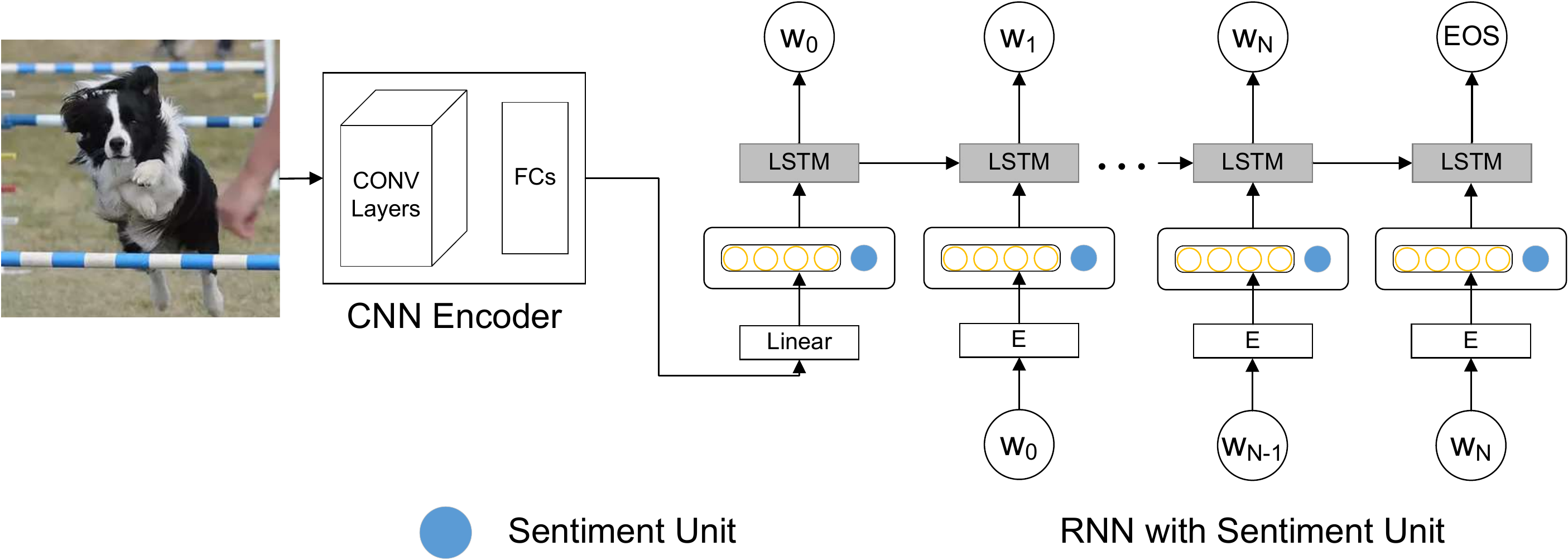}
	\caption{Direct injection of sentiment. We concatenate the sentiment label with the output of the embedding layer, which is the input of the Recurrent Neural Network. The value of sentiment unit can only be -1, 0 or 1, which represents negative, neutral or positive sentiment.}
	\label{fig:direct}
	\vspace{-10pt}
\end{figure}
\subsection{Direct Injection}
Recently, a sentiment unit is discovered in a RNN based generation model~\cite{radford2017learning}. After careful analysis and inspection, the authors find a single unit within the RNN model, which as their results suggest, is directly related to sentiment. The value distributions of different sentiments are quite different. This result suggests that neural networks can learn the sentiment semantic meaning and more importantly, this semantic meaning can be controlled by one unit. 

Inspired by the above finding, we design a clean way of injecting sentiment to the current image captioning system. We implement our captioning system on top of the neural caption generator from~\cite{vinyals2015show}, which is one of the most influential and effective algorithms on the MS-COCO Image Captioning Challenge. In this model, each word is passed through an embedding layer, which produces the input for the recurrent neural network. Our design is to add the sentiment unit dimension before passing the output to the recurrent unit. We name this approach direct injection. The sentiment information is represented as a single dimension.

As shown in \figurename~\ref{fig:direct}, sentiment unit is the additional dimension concatenated to the input of the recurrent unit. The value of the sentiment unit can only be -1, 0 or 1, which represents negative, neutral or positive sentiment. When training the model, this unit is set to the sentiment label of current input caption. More interestingly, in the testing stage, we can manipulate the value of this unit to generate different captions with the corresponding sentiments. 

In our implementation, we employ the Long-Short Term Memory (LSTM) cell to implement the recurrent unit. The input to the LSTM cell is being further processed to produce its output. In the LSTM cell, there are input gate, output gate, forget gate and memory gate. Take input gate for example, the output of this gate is computed as follows:
\begin{equation}
i_t = \sigma_g(W_i x_t + U_i h_{t-1} ),
\end{equation}
where $W_i$ and $U_i$ are the parameters, $x_t$ is current input and $h_{t-1}$ is previous hidden state. When we add the sentiment unit to the input, we only concatenate the sentiment value to $x_t$. When the input caption is neutral, the value of this additional position will be set to $0$ and thus the sentiment unit makes no contribution to the output of the input gate. Still, the remaining positions of the recurrent unit model the language. On the other hand, positive sentiment and negative sentiment make opposite contribution to the output. They provide additional sentiment information besides language modeling. This intuitively explains the motivation to choose the value of -1, 0 and 1 to represent different sentiment. Without loss of generality, this analysis is also valid for other types of recurrent unit.

\subsubsection{Sentiment Loss}
With the additional sentiment unit, we expect that the hidden state from the LSTM cell should also be able to distinguish between different sentiments. In particular, we include the sentiment loss on top of the hidden state of the recurrent unit. The loss is the negative log-likelihood of the ground-truth sentiment. We define it as
\begin{equation}
L(h_t, l) = -\sum_{j} I(j = l)\log(p(l= j|h_t)),
\end{equation} 
where $I(\cdot)$ is an indicator function and $p(i^k=j|h_t)$ is the output of a multilayer perceptron (MLP). The MLP takes the current hidden state of the RNN $h_t$ as the input and produces the probability of assigning the current state to the three sentiment classes.

The loss is computed at every step of the RNN and averaged over all steps during the training stage. In other words, the context of the RNN (hidden state) is not only used to predict the next word, but also used to predict the current sentiment state. 
\subsection{Injection by Sentiment Flow}
\label{sec:sf}
Direct injection in the previous section adds a sentiment unit at every step of the recurrent neural network. In such a way, the model is less likely to ignore the sentiment signal and thus will take this into consideration in learning the parameters. However, sometimes, this kind of \textit{hard} sentiment signal could also be problematic because most of the words or phrases are irrelevant to sentiment. Meanwhile, these words or phrases are shared by many captions belonging to different sentiment groups. For example, the following two training captions for the same image have opposite sentiments.
\begin{enumerate}\small
	\item \textit{an elephant is standing over by some amazing trees}.
	\item \textit{an elephant is standing over by some scary trees}.
\end{enumerate}
When training the models using the above two captions, we need to provide totally opposite sentiment labels for all the words "\textit{an elephant is standing over by some}" in the direct injection approach. However, they are essentially in the same state. Only when the system sees "amazing" or "scary", it starts to go to different states with different sentiments. 

The above observation suggests that direct injection could potentially lead to confusions of the model due to its mechanism of  providing the sentiment information. A possible solution is to allow the model itself to decide, at each step, whether to use the sentiment information or not. Indeed, this is quite similar to the philosophy of a memory cell in the LSTM network, which tries to preserve information and propagate the information over the sequence.

Inspired by this analogy, we design and implement the sentiment cell, which has a similar computation mechanism as the memory cell. At each step, the sentiment cell tries to update its state by looking at the states of the input gate, the forget gate and the output gate. Similarly, the updated sentiment state is also employed to produce the current hidden state, which is then employed to predict the next context word in the sequence. Differently, the sentiment state is initialized by the sentiment label of the current captions, where the memory cell is usually initialized as an all-zero vector. In such a way, we expect that the sentiment information, provided at the initial sentiment unit, will flow over the sequence and be updated at each step. Later, the propagated sentiment information allows the hidden state to accurately predict sentiment related next context word.

\begin{figure}[!h]
		\centering
		\includegraphics[width=0.45\textwidth]{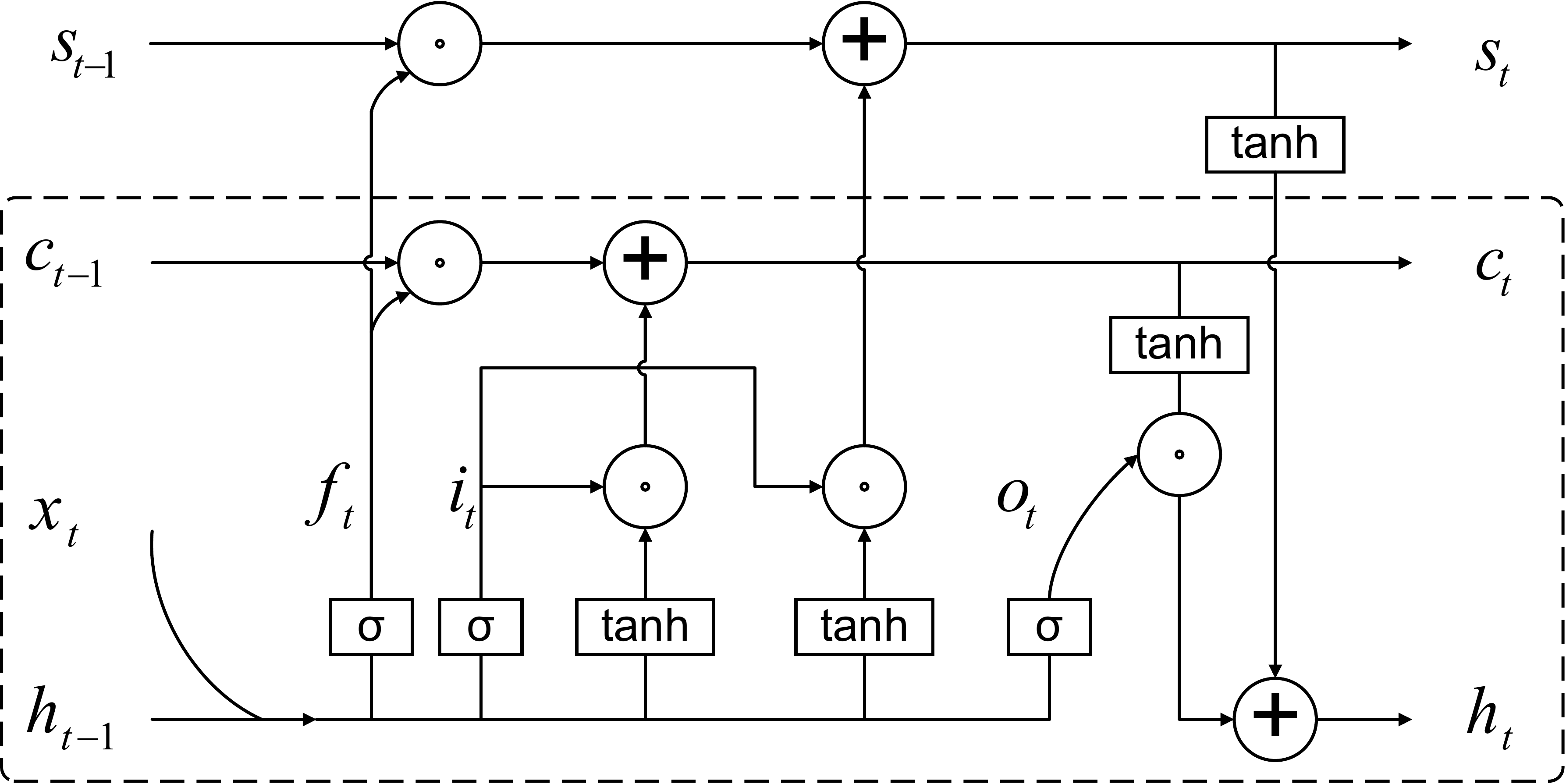}
		\caption{Proposed LSTM with A Sentiment Cell. The sentiment cell $s$ is similar to the memory cell $c$, where information is propagated over the sequence. The module inside the dashed line box is the LSTM unit without sentiment cell.}\label{fig:lstms}
		\vspace{-10pt}
\end{figure}

\figurename~\ref{fig:lstms} shows the comparisons between the traditional LSTM cell and the proposed LSTM cell with a sentiment cell. Similar to the memory cell, the information of sentiment cell is propagated over the sequence and is updated at each step. We compute the sentiment cell as follows
\begin{equation}
s_t = f_t \odot s_{t-1} + i _t \odot \sigma(W_{x}^s x_t + W^s_h h_{t-1} + b_s), \label{eqn:st}
\end{equation}
where $f_t$ and $i_t$ are the current states of forget gate and input gate, respectively.\footnote{Since LSTM is well known in the community, we do not include the equations for computing the states of different gates.} Accordingly, the hidden state also takes the current state of the sentiment cell into consideration to predict the next context word:  
\begin{equation}
h_t = o_t \odot (\sigma_h(c_t) + \sigma_h(s_t))
\end{equation}
where $\sigma_h$ is the activation function, $o_t$ is the state of the output gate. $c_t$ and $s_t$ are the current states for the memory cell and sentiment cell, respectively. The sentiment signal is provided at the initial step of the recurrent network. Specifically, let $l$ be the label for the current caption, we then compute the initial sentiment cell state $s_0$ as 
\begin{equation}
s_0 = \sigma_s(W_s E(l) + b),
\end{equation}
where $\sigma_s$ is the activation function, $b$ is the bias term and $E$ is the embedding layer, which maps the  sentiment label to a vector. We choose $\tanh$ as the activation function for $\sigma_s$ to make the value ranges of $s_0$ consistent with the subsequent $s_i$ ($i > 0$). 

So far, we have discussed the mechanism to propagate the sentiment information over the sequence. The learning of the parameters related to the sentiment unit can be learned by back propagating the error in predicting the next word in the sequence. Since the sentiment signal is only provided at the initial step of the recurrent network, this could lead to the diminishing of the sentiment signal over the sequence. To circumvent this problem, we apply the same sentiment loss as in the direct injection case to help the model maintain the sentiment signal in the propagation process. Again, we include the sentiment loss as part of our training objective. 

However, different from the direct injection, where sentiment losses are computed at each individual step of the recurrent network, here we only compute the sentiment losses at the end of each training instance. In such a way, we expect that the sentiment signal should be maintained and distinguishable between different sentiment classes. Specifically, we compute the sentiment loss for each input pair of input sentence $x$ and sentiment label $l$ as follows
\begin{equation}
L(x, l) = -\sum_{j} I(j = l)\log(p(\hat{l}_{|x|}= j|s_{|x|})),
\end{equation}
where $s^k$ is the $k$-th training caption, $l^k$ is its sentiment label and $\hat{l}_{|x^k|}$ is the predicted sentiment label at the end of the caption using the last sentiment unit state $s_{|x|}$.
\section{Experiments}
\begin{table*}[!h]
	\centering
	\small
	\begin{tabular}{*{9}{c}}
		\hline
		Testing Set & Models & B-1 & B-2 & B-3 & B-4 & ROUGE-L & METEOR & CIDEr\\
		\hline
		\multirow{4}{*}{POS} &  CNN+RNN    &    0.487    &    0.281    &    0.170    &    0.107    &    0.366    &    0.153    &    0.556    \\ 
		&SentiCap    &    0.491    &    0.291    &    0.175    &    0.108    &    0.365    &    0.168    &    0.544    \\
		&Direct Injection (\textbf{Ours})    &    \textbf{0.512}    &    0.306    &    0.188    &    0.116     &    0.384    &    \textbf{0.172}    &    \textbf{0.611}    \\
		&Sentiment Flow (\textbf{Ours})    &    0.511    &    \textbf{0.314}    &    \textbf{0.194}    &    \textbf{0.123}&    \textbf{0.386}        &    0.169    &    0.608    \\
		\hline
		\multirow{4}{*}{NEG} &    CNN+RNN    &    0.476    &    0.275    &    0.163    &    0.098    &    0.361    &    0.150    &    0.546    \\
		&    SentiCap    &    0.500    &    0.312    &    0.203    &    0.131    &    0.379    &    0.168    &    0.618    \\
		&    Direct Injection (\textbf{Ours})    &    \textbf{0.522}    &    \textbf{0.336}    &    \textbf{0.222}    &    0.146    &    \textbf{0.398}    &    \textbf{0.171}    &    0.684    \\
		&    Sentiment Flow (\textbf{Ours})    &    0.510    &    0.330    &    0.219    &    \textbf{0.148}    &    0.394    &    0.170    &    \textbf{0.701}    \\
		\hline
	\end{tabular}
	\caption{Comparisons of different approaches in terms of benchmark metrics. }
\label{tab:results}
\vspace{-10pt}
\end{table*}
\begin{table*}[!ht]
	\centering
	\small
	\begin{tabular}{*{9}{c}}
		\hline
		Testing Set & Models & B-1 & B-2 & B-3 & B-4 & ROUGE-L & METEOR & CIDEr\\
		\hline
		\multirow{4}{*}{POS} & Direct Injection    &    0.51    &    \textbf{0.312}    &    \textbf{0.191}    &    \textbf{0.117}    &    \textbf{0.385}    &    \textbf{0.175}    &    \textbf{0.614}    \\
		&Direct Injection (S-Loss)    &    \textbf{0.512}    &    0.306    &    0.188    &    0.116     &    0.384    &    0.172    &    0.611    \\
		\cline{2-9}
		&Sentiment Flow    &    \textbf{0.513}    &    0.31    &    0.189    &    0.115 &    \textbf{0.388}        &    \textbf{0.171}    &    \textbf{0.621} \\
		&Sentiment Flow (S-Loss)&    0.511    &    \textbf{0.314}    &    \textbf{0.194}    &    \textbf{0.123}&    0.386        &    0.169    &    0.608   \\    
		\hline
		\multirow{4}{*}{NEG} &    Direct Injection    &    0.483    &    0.298    &    0.191 & 0.124    &    0.376 & \textbf{0.171} & 0.584    \\
		&    Direct Injection (S-Loss)    &    \textbf{0.522}    &    \textbf{0.336}    &    \textbf{0.222}    &    \textbf{0.146}    &    \textbf{0.398}    &    \textbf{0.171}    &    \textbf{0.684}    \\
		\cline{2-9}
		&    Sentiment Flow    &    0.504    &    0.313    &    0.199    &    0.129    &    0.387    &    0.167    &    0.643    \\
		&    Sentiment Flow (S-Loss)    &    \textbf{0.51}    &    \textbf{0.33}    &    \textbf{0.219}    &    \textbf{0.148}    &    \textbf{0.394}    &    \textbf{0.17}    &    \textbf{0.701}    \\
		\hline
		\multirow{4}{*}{AVG} &    Direct Injection &    0.4965    &    0.305    &    0.191    &    0.1205    &    0.3805    &    \textbf{0.173}    &    0.599    \\ 
		&    Direct Injection (S-Loss)    & \textbf{0.517}    &    \textbf{0.321}    &    \textbf{0.205}    &    \textbf{0.131}    &    \textbf{0.391}    &    0.1715    &    \textbf{0.6475}    \\
		\cline{2-9}                                                        
		&    Sentiment Flow&    0.5085    &    0.3115    &    0.194    &    0.122    &    0.3875    &    0.169    &    0.632    \\
		&    Sentiment Flow (S-Loss)        &    \textbf{0.5105}    &    \textbf{0.322}    &\textbf{0.2065}    &    \textbf{0.1355}    &    \textbf{0.39}    &    \textbf{0.1695}    &    \textbf{0.6545} \\
		\hline
	\end{tabular}
	\caption{Impact of the sentiment loss (L-LOSS) on the performance of the proposed models. AVG represents the averaged performance of the POS and NEG sets, from which we can obtain the overall approximate impact of this loss term. }
\label{tab:loss_results}
\vspace{-10pt}
\end{table*}

We implement and evaluate the above two approaches. The source codes along with the trained model will be publicly available upon the acceptance of this work.
\subsection{Dataset}
For the image captioning task, MS-COCO dataset is the largest publicly available and manually labeled dataset, which is a valuable and high quality dataset. However, there are no sentimental captions in this dataset. Usually, there are two approaches to building the dataset. The authors in~\cite{mathews2016senticap} construct a manually labeled sentiment dataset, SentiCap. They conduct a caption re-writing task using the Amazon Mechanical Turk (AMT). In particular, the task asks AMT workers to re-write the captions of some subset of the MS-COCO images by incorporating sentiment related words or phrases. In this way, the collected captions are high quality and in the same domain as the MS-COCO dataset. Indeed, the state-of-the-art approach in~\cite{mathews2016senticap} employed MS-COCO for pre-training a captioning system. They then learn the sentiment related component on their small but high-quality sentiment related caption dataset. \tablename~\ref{tab:dataset} summarizes the statistics of the two manually constructed datasets. MS-COCO is much larger than SentiCap in~\cite{mathews2016senticap}.
\begin{table}[!h]
	\small
	\begin{tabular}{*{5}{c}}
		\hline
		Dataset & Statistics & Training &Testing &Validating \\
		\hline
		\multirow{2}{*}{MS-COCO}  & \# of Images & 82,783 & 40,775   & 40,504 \\
		& \# of captions & 414,113 & NA & 202,654 \\
		\hline
		\multirow{2}{*}{SentiCap Pos}  & \# of Images & 824  & 673  & 174 \\
		& \# of captions & 2,380& 2,019 & 409  \\
		\hline
		\multirow{2}{*}{SentiCap Neg}  & \# of Images & 823 & 503  & 174 \\
		& \# of captions & 2,039 & 1,509 & 429 \\
		\hline
	\end{tabular}
	\caption{Statistics of the two image captioning datasets constructed by crowd sourcing.}
\label{tab:dataset}
\vspace{-10pt}
\end{table}

The second approach is to collect a weak captioning dataset. The study in~\cite{ordonez2011im2text} collects a total of one million image and caption pairs, where the captions are composed by Flickr users. The approach is much cheaper and can build a very large dataset. But the collected captions are noisy and low-quality compared with the manually labeled approach. More recently, a large weak sentiment caption dataset is collected in~\cite{shinimage}. 

Current studies on image captioning can also be grouped into two categories. The first group uses a high-quality image dataset. The goal is try to design a better model to understand both visual and textual content to produce human-like captions. The second group of studies employs the weak labeled captions. One of the goals is to generate reasonable captions. Meanwhile, it needs to overcome the noise in the dataset. Our work belongs to the first group. Therefore, we evaluate and compare the proposed approaches with the state-of-the-art approaches on the high-quality manually labeled dataset.
\subsection{Implementation Details}
We implement the proposed approach using the PyTorch (\url{http://pytorch.org}) neural network library due to its high performance compared with other deep learning libraries. All the experiments are run on a Linux server with two NVIDIA Titan X GPUs. For the CNN encoder shown in \figurename~\ref{fig:direct}, we use the pretrained ResNet-152~\cite{he2016deep} to extract visual features. We also compare the results of ResNet with pretrained VGG-16~\cite{simonyan2014very}. It turns out they have very comparable performance. However, VGG is slow and consume more computational resources than ResNet. Thus, in the following experiments, we choose to use ResNet.

In our implementation, the size of the embedding layer is $256$ and the hidden size of the recurrent network is $512$, which are common choices for image captioning systems. We train the model using Adam~\cite{kingma2014adam} optimization algorithm with an initial learning rate of $0.001$. The size of the mini-batch is $150$. The loss over each mini-batch is averaged and back propagated to all the parameters in our model. However, we keep the parameters of ResNet fixed during the training, which can speed up the training but may give up some performance.

During the testing stage, when given a new image and the sentiment label, we use the ResNet and the sentiment label to initialize the states of the recurrent network. Next, at each step, we use beam search to generate several candidate captions and the caption with the smallest loss is returned as the final caption for the given testing image.
\subsubsection{Training Strategy}
As discussed in the previous section, we focus on using the manually labeled dataset for producing human-like captions. In~\cite{mathews2016senticap}, the authors proposed a switching RNN model for captions with sentiments. They first employ the MS-COCO data to pre-train a RNN, which mainly controls the generation of the sentences. Next, a second RNN network is trained using the smaller dataset, which mainly controls the generation of sentiments. Similarly, we also train our model using both the MS-COCO and the SentiCap datasets. However, our strategy is simple yet effective. We assign \textit{neutral} to all the captions in MS-COCO dataset. Next, we combine all the training sets of SentiCap and MS-COCO as our training dataset. This training dataset is highly unbalanced due to the large number of neural captions. However, the main contribution from MS-COCO is to help the learning of RNN for generating sentences and learn the semantic mapping between images and sentences. The small number of positive and negative captions help the learning of sentiment related parameters.

\subsubsection{Models for Comparison}
The state-of-the-art approach on the task of generating captions with sentiments is from~\cite{mathews2016senticap}. They proposed a switching RNN for this task. Overall, their model achieved the best performance compared with other baselines. However, their model cannot be trained in an end-to-end fashion. For comparison, we also include the results of the model~\cite{vinyals2015show}, from which we have borrowed the main framework for our captioning system. This model is also included as one of the baselines in~\cite{mathews2016senticap}.
\subsection{Experimental Results}
We train and evaluate the proposed two models with the same dataset splits of the two baselines. When generating the captions, our trained model is provided with the image and the sentiment label. The results on the testing split of the captions with sentiments are shown in \tablename~\ref{tab:results}. We use the MS COCO evaluation tool\footnote{\url{https://github.com/tylin/coco-caption}} to obtain all the performance results using the generated captions. Overall, the proposed two models outperform the two baselines over all the evaluation metrics. The performances of the two proposed approaches are comparable. 

Meanwhile, it is also interesting to note that the sentiment flow approach seems to have a balanced performance on both the POS and the NEG sets. For the direct injection approach, the performance on the NEG set seems to be better than the POS category. This is also consistent with the results of the SentiCap. Part of the reason can be attributed to the larger diversity of the POS testing set.\footnote{According to~\cite{mathews2016senticap}, they use 1,027 positive adjective noun pairs as candidates to construct the positive captions. Comparably, there are only 436 negative adjective pairs.}

\subsection{Impact of Sentiment Loss}
Both direct injection and sentiment flow employ the sentiment loss to help the learning process. Direct injection applies the sentiment loss at each individual step of the recurrent network. Sentiment flow only computes the loss at the last step. In this section, we analyze the impact of this loss term on the performance of caption generation. 

For each of the proposed model, we also train another model, which uses the same configuration (hyper-parameters, initializations and so on) but without the sentiment loss during back propagation. We also use the same approach for generating the candidate captions for each testing image. 

\tablename~\ref{tab:loss_results} shows the performance of different models trained with and without the sentiment loss term. For the sentiment flow model, this loss helps improve the performance of the model on most of the evaluation metrics. For the direct injection model, the sentiment loss brings significant performance improvements in the \textit{negative} testing set. However, it leads to worse performance on the \textit{positive} testing set. This could be partially because we have provided the sentiment signal at every step of the recurrent network, which implies that the sentiment signal is already strong enough. Therefore, the sentiment loss is unnecessary to ensure the propagation of the sentiment signal. Still, if we average the performance for both the \textit{positive} and \textit{negative} sets, it suggests that the sentiment loss could bring performance improvements for the two proposed models.

\begin{figure}[!ht]
	\centering
	\begin{subfigure}[b]{0.45\textwidth}
		\centering
		\includegraphics[width=\textwidth]{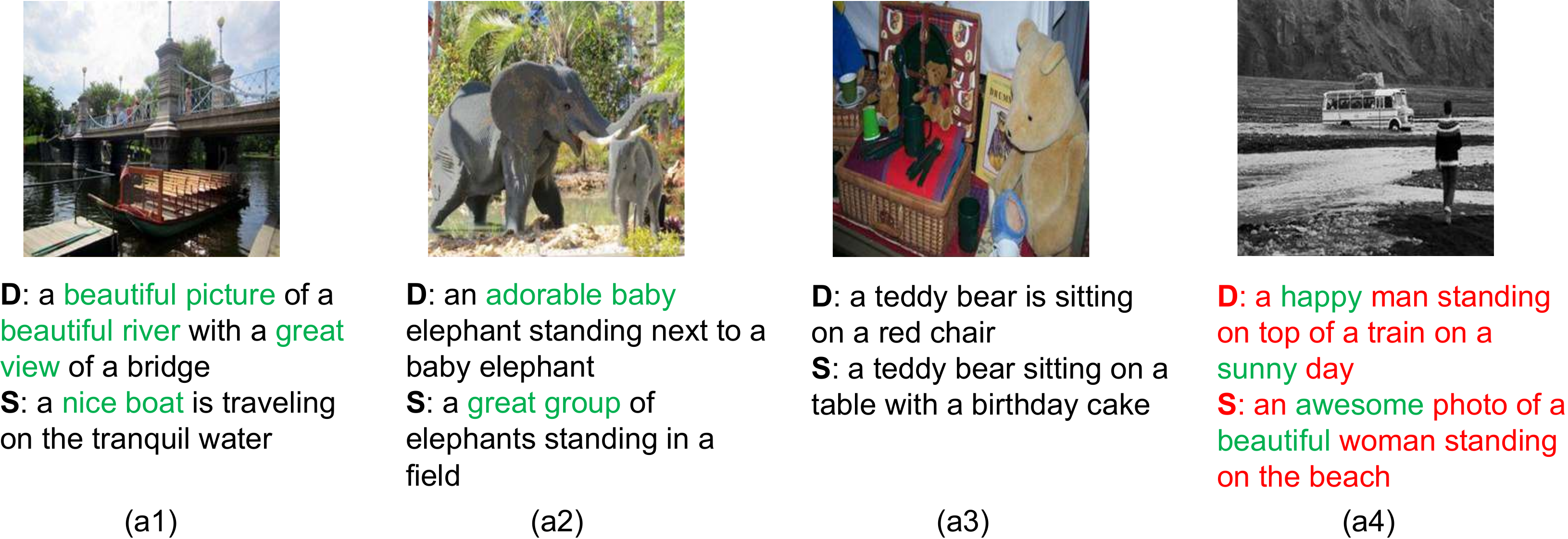}
		\caption{Examples from the Positive Testing Set.}
		\label{fig:example:pos}
		\vspace{10pt}
	\end{subfigure}
	
	\begin{subfigure}[b]{0.45\textwidth}
		\includegraphics[width=\textwidth]{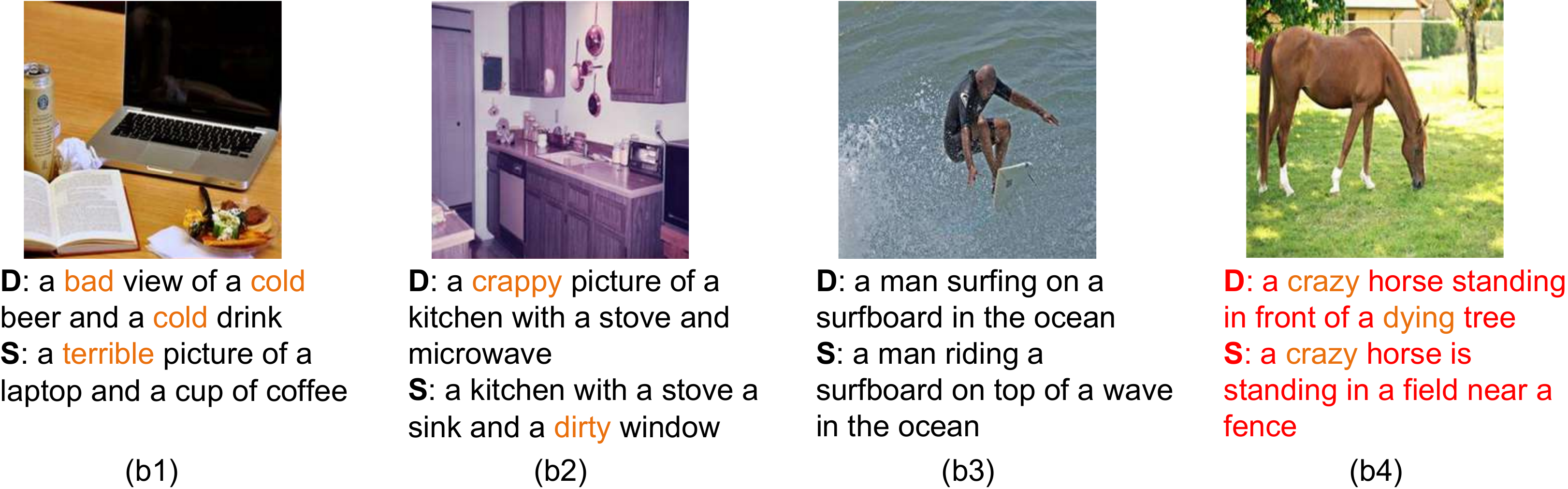}
		\caption{Examples from the Negative Testing Set.}
		\label{fig:example:neg}
	\end{subfigure}
	\caption{Examples of the generated captions (\textbf{D} for direct injection and \textbf{S} for sentiment flow). Positive and negative  words are highlighted in green and orange color, respectively.}
	\label{fig:example}
\end{figure}
\subsection{Examples of Generated Captions by Different Models}
\figurename~\ref{fig:example} shows some examples of the generated captions for both proposed models. \figurename~\ref{fig:example:pos} contains positive caption examples. The sentiment related words are highlighted in green color. From (a1)-(a2), we can discover that positive sentiment related words, such as \textit{beautiful, nice, adorable}, are injected into the generated captions, which also match the content of the corresponding images.  There are examples, where both models return factual descriptions. (a3) shows one such example. For (a4), the sentiment of the given image in the positive testing set is likely to be negative due to its gray color. Indeed, the three ground truth positive captions for this given image are\\
{
	\footnotesize
	1. \textit{a nice man that is standing in the dirt}\\
	2. \textit{a nice person walks near water and there is a bus driving through}\\
	3. \textit{a bus filled with wonderful people and luggage is crossing a small stream.}\\
}
For the workers, it is also challenging to re-write the captions for this given image. The sentiment related words are general and are not specifically related to the content of the given image.

\figurename~\ref{fig:example:neg} contains negative caption examples. Similarly, (b1)-(b2) present examples using negative related words such as \textit{bad, terrible and crappy}. For (b4), the overall sentiment of the image should be positive. Therefore, the produced negative caption may be inconsistent with the content of the image. The ground truth negative captions for this given image are\\
{
\footnotesize
1. \textit{a crazy horse with white socks grazing in the grass near a home}\\
2.  \textit{that crazy horse is eating poisonous mushrooms where is his owner}\\
3. \textit{the crazy horse is standing alone on the grass in the yard}.
}
	
However, \textit{crazy} may not be correct to describe this horse and \textit{poisonous mushrooms} are not shown in the image. This, on the other hand, also indicates that even for humans, it is quite challenging to write sentimental captions for some images.

\begin{figure}[!ht]
	\centering
	\includegraphics[width=.45\textwidth]{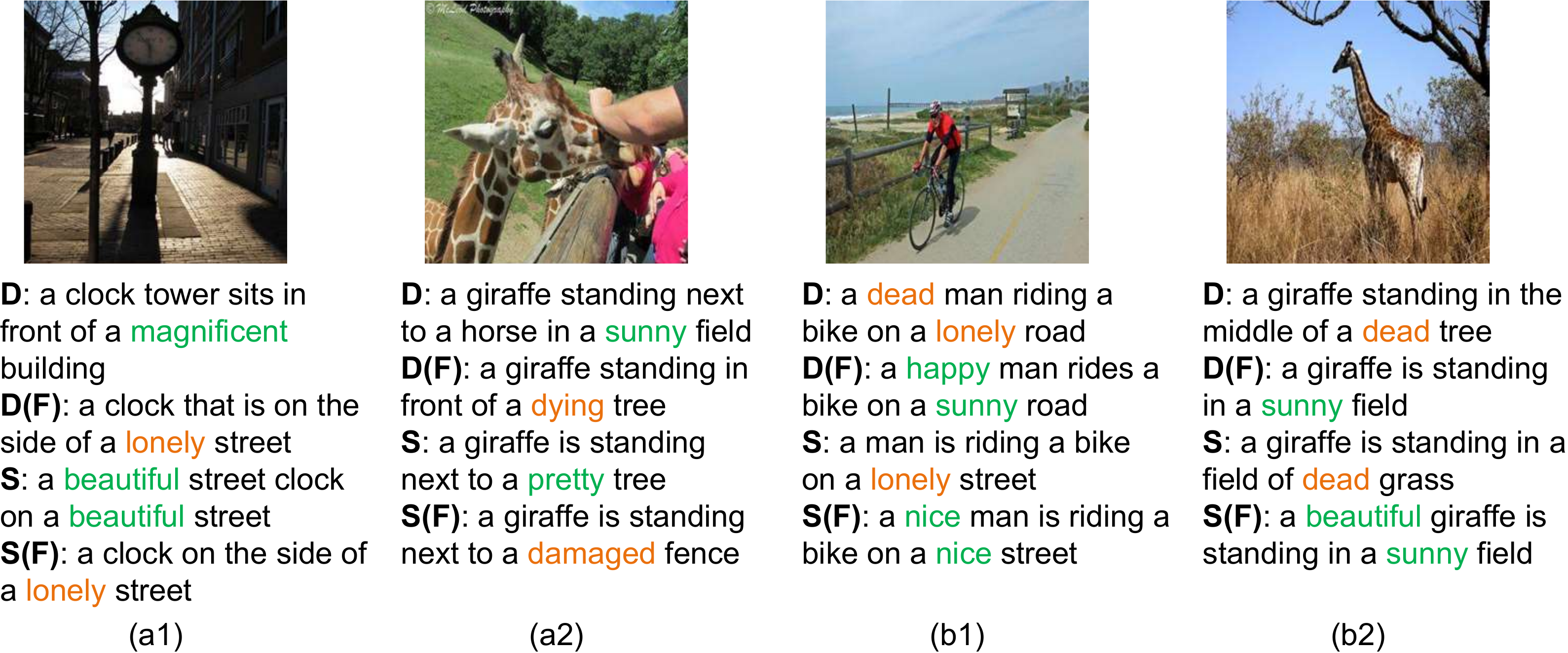}
	\caption{Examples of the generated captions (\textbf{D} for direct injection and \textbf{S} for sentiment flow). Positive and negative  words are highlighted in green and orange color, respectively.}
	\label{fig:flip}
	\vspace{-10pt}
\end{figure}

\subsection{Controllable Sentiment Caption Generation}
Because in both proposed models, we only use the sentiment label to control the input of the sentiment signal. Therefore, we can easily manipulate the generation of sentimental captions by simply flipping the label of the sentiment. In this section, we show some analysis of the results on controlling the sentiment of the given image.

\begin{table}[!h]
	\small
	\begin{tabular}{*{5}{c}}
		\hline
		Model & Total & Matched &Total (F) & Matched (F) \\
		\hline
		D-POS & 88.9\% & 82.8\% (POS) & 89.2\%   & 81.9\% (NEG)\\
		S-POS & 83.8\% & 77.7\% (POS)& 75.9\% & 58.8\% (NEG)\\
		D-NEG & 91.1\% & 84.5\% (NEG)& 85.7\% & 75.5\% (POS)\\
		S-NEG & 80.3\% & 62.8\% (NEG)& 83.7\% & 76.3\% (POS)\\
		\hline
	\end{tabular}
	\caption{Percentages of sentiment captions. \textit{Total} column is the percentage of sentiment captions. The \textit{Matched} column is the percentage of captions, where their sentiments match the sentiment labels of the testing set. The last two columns are the results after we flip the sentiment labels.}
\label{tab:flip}
\end{table}
\figurename~\ref{fig:flip} shows some examples of the generated captions of the two proposed models. For each example, we show both the caption produced by providing the original sentiment label and the caption after we flip the sentiment label. Positive and negative sentiment words are highlighted in green and orange colors, respectively. After we simply change the input of the sentiment signal, the model intends to inject matching sentiment words onto different objects in the given image. For example, in (b2), both the models focus on the tree or grass when providing negative sentiment (original sentiment label). However, when we change the sentiment label to positive, they inject positive words to other objects (giraffe and field).

We summarize the percentages of sentimental captions. For each generated caption, if it has one of the sentiment words\footnote{See~\cite{mathews2016senticap} for details of the sentiment words.}, we will count it as a sentimental caption. Furthermore, if the label of the sentiment word is the same as the given sentiment label for the generation of the caption, we count it as the matched sentiment caption. \tablename~\ref{tab:flip} shows the results. The last two columns are the percentages after we flip the sentiment labels. Overall, we can see that direct injection can produce a higher percentage of sentiment captions. The sentiment signal may become weak in the sentiment flow model. Especially, for the generation of negative captions, the sentiment flow model seems to have the lowest percentage of sentiment captions.

Overall, both models can easily generate captions with different sentiment labels. Interestingly, the model may inject sentiment words, which match the given sentiment label, into the produced captions. However, the stepwise direct injection of sentiment seems to have better performance regarding preserving the sentiment signal and thus provides matching sentiment captions.
\section{Conclusions}
In this work, we present two different approaches for injecting sentiment into the generation of captions. Our  models outperform the state-of-the-art approaches on the manually constructed image captioning datasets. The proposed model is simple and easy to deploy. We also study the impact of manipulating the sentiment labels. The results suggest that the model can learn the semantic mapping between images and captions. At the same time, it can also produce captions with the supplied sentiment labels. 

At present, the proposed model can only be trained with the manually labeled captions. Future work can focus on how to develop effective approaches to produce different styles of captions by learing from large-scale weakly labeled data. Additionally, we also notice that sometimes if the sentiment label does not match the human perceived sentiment from the content of the image, the system may not produce semantically correct captions. Therefore, another possible future research direction is to automatically learn the additional attributes from the image and then produce captions with matching attributes. For example, the recent methods on analyzing visual sentiment~\cite{borth2013large,you2015PCNN} could be employed for automatically detecting image sentiment first and then generating the matching sentiment captions.

\bibliography{aaai}
\bibliographystyle{aaai}
\end{document}